# BioFace3D: A fully automatic pipeline for facial biomarkers extraction of 3D face reconstructions segmented from MRI

Álvaro Heredia-Lidón[1], Luis M. Echeverry-Quiceno[2], Alejandro González[1], Noemí Hostalet[2,3,4], Edith Pomarol-Clotet[3,4], Juan Fortea[5], Mar Fatjó-Vilas[2,3,4], Neus Martínez-Abadías[2], Xavier Sevillano[1]

**1** HER - Human-Environment Research Group, La Salle - Universitat Ramon Llull, Barcelona, Spain
**2** Departament de Biologia Evolutiva, Ecologia i Ciències Ambientals (BEECA), Facultat de Biologia, Universitat de Barcelona (UB), Barcelona, Spain
**3** FIDMAG, Sisters Hospitallers Research Foundation, Barcelona, Spain
**4** CIBERSAM (Biomedical Research Network in Mental Health, Instituto de Salud Carlos III), Madrid, Spain
**5** Sant Pau Memory Unit, Hospital de Sant Pau i la Santa Creu, Barcelona, Spain

* alvaro.heredia@salle.url.edu

## Abstract

Facial dysmorphologies have emerged as potential critical indicators in the diagnosis and prognosis of genetic, psychotic and rare disorders. While in certain conditions these dysmorphologies are severe, in other cases may be subtle and not perceivable to the human eye, requiring precise quantitative tools for their identification. Manual coding of facial dysmorphologies is a burdensome task and is subject to inter- and intra-observer variability. To overcome this gap, we present BioFace3D as a fully automatic tool for the calculation of facial biomarkers using facial models reconstructed from magnetic resonance images. The tool is divided into three automatic modules for the extraction of 3D facial models from magnetic resonance images, the registration of homologous 3D landmarks encoding facial morphology, and the calculation of facial biomarkers from anatomical landmarks coordinates using geometric morphometrics techniques.

## Author summary

Facial dysmorphologies emerge as an indicator of genetic, psychotic and rare disorders. The study of these dsymorphologies requires accurate and automated quantitative tools. Therefore, we present BioFace3D, an automatic tool to calculate facial biomarkers from 3D facial models reconstructed from magnetic resonance imaging (MRI). BioFace3D includes three modules: extraction of 3D facial models from MRI, registration of landmarks to capture facial morphology, and calculation of facial biomarkers using geometric morphometrics.

## Introduction

Accumulating evidence indicates that facial dysmorphologies are associated with multiple genetic, psychotic, and rare disorders with a neurodevelopmental component, appearing as accessible and promising diagnostic factors [1–6]. Each condition with associated facial dysmorphologies presents a distinctive facial pattern that is unique, correlated with brain alterations, and often sex-specific [6]. In some cases, facial dysmorphologies are severe and readily detectable, such as in Down or Fragile-X syndromes [7, 8]. In other conditions, like in autism [9] or in psychotic disorders [10, 11], they are much subtler, of little or no functional or cosmetic consequence and indistinguishable to the untrained eye [12].

In either case, the high prevalence of facial dysmorphologies has led to a growing interest to obtain facial biomarkers that can be leveraged with diagnosis and prognosis purposes [13]. In this context, face image-based diagnostic tools like Face2Gene [14] have been widely adopted in clinical practice. However, this tool often requires additional clinical patient information to increase diagnostic accuracy [15] and it disregards 3D facial shape information, thus possibly introducing significant distortion of apparent facial morphology [2].

For this reason, healthcare professionals still use qualitative visual assessments and simple anthropometric measurements to guide diagnosis. To optimize this process, quantitative and objective automatic tools for assessing 3D facial dysmorphologies are needed.

## Geometric morphometrics and 3D imaging to quantify facial shape

One of the most powerful approaches to accurately quantify facial shape, and consequently, to obtain facial biomarkers of potential application in clinical practice is the combination of geometric morphometrics with 3D facial imaging [2].

Geometric morphometrics (GM) is a sophisticated body of robust statistical tools developed for measuring and comparing 3D shapes with increased precision and efficiency [16, 17]. GM analyses are based on calculations on the Cartesian coordinates of anatomical landmarks [18, 19]. In our case, biologically homologous points that could be reliably recorded in all individuals were registered on the surface of a three-dimensional model of the face.

Different GM techniques are commonly used to assess shape variations and to detect differences between groups: *i)* Generalized Procrustes Analysis (GPA) [20], which projects the landmarks coordinates of the individuals onto a common morphospace for global shape analysis [21], and *ii)* Euclidean Distance Matrix Analysis (EDMA) [22], which searches for statistically significant local shape differences among linear distances between landmarks. The use of both GPA and EDMA is complementary and allows to compute facial biomarkers that represent facial shape both globally and locally.

These biomarkers can be calculated on 3D facial models commonly acquired by suitable 3D imaging techniques, like 3D stereophotogrammetry [23], laser scans [24], structured light cameras [2, 25], or deep learning-based 3D face reconstruction from several 2D images [26, 27]. Moreover, facial meshes can also be extracted by segmentation of medical image modalities that usually target brain imaging, like computer tomography (CT) and magnetic resonance images (MRI) of the head, with an adaptation of the scanning protocols to ensure the capture of the whole face surface [28].

## Towards an automatic pipeline for facial biomarkers extraction from MRI

Recent studies have highlighted the correlation between brain and facial morphologies in genetic and neurodevelopmental pathologies, such as psychotic disorders [6]. Since both the face and the brain derive from ectodermal derivatives and intimately develop in space and time through regulation of common signaling pathways, factors altering brain development can also impact facial development [29] and cause characteristic craniofacial phenotypes [30].

Therefore, full head MRI scans emerge as a valuable imaging source to simultaneously study brain and face morphologies. However, while there exist multiple software tools that automate several neuroimaging analysis tasks, like SPM12 [31] or FreeSurfer [32], there is no analogue for automatic facial shape morphology analysis from MRI.

The identification of facial biomarkers of diagnostic potential requires processing large volumes of MRI scans, entailing *i)* the reconstruction of 3D facial models from MRI, *ii)* the normalization of data for subsequent analysis, and *iii)* the registration of anatomical landmarks on the 3D facial models. Tackling all these tasks manually is unfeasible, not only due to their burdensomeness but also because they require deep anatomical expertise and are subject to inter- and intra-observer variability [33].

For this reason, we introduce BioFace3D, an automatic high-throughput pipeline for computing facial biomarkers from full head MRI scans. BioFace3D automatically extracts 3D facial isosurfaces from MRI, registers 3D facial landmarks on the extracted meshes, and computes GM-based facial biomarkers. This allows conducting morphometric studies to identify local and global facial shape features related to any condition with associated facial dysmorphologies. To our knowledge, this is the first end-to-end tool publicly available for the advanced study of facial morphology extracting facial biomarkers from MRI scan data.

## Materials and methods

### Datasets

**Training datasets**

Training the algorithms in BioFace3D required the manual processing and annotation of a proprietary dataset of head MRI scans. This dataset included 538 high-resolution T1-weighted (T1w) head MRI scans, provided by FIDMAG Germanes Hospitalàries Research Foundation from Hospital Sant Joan de Déu (1.5 GE Sigma scanner) and Fundació Pasqual Maragall (Siemens Magnetom PRISMA 3T), and Hospital Sant Pau Memory Unit (Philips 3 Tesla X Series Achieva), all from Barcelona, Spain. Each participant was scanned at only one of these centres. The study adhered to the guidelines set by the participating institutions and was approved by their respective ethics committees (PR-2023–21, IRB00003099, CER122317 and CER URL 2023 2024 012). All procedures were carried out according to the Declaration of Helsinki. A summary of the dataset composition is presented in S1 Table.

From the NIfTI files derived from each MRI scan, 3D head reconstructions were manually generated using the 3D analysis software for scientific data Amira 2019.2 (Thermo Fisher Scientific, Waltham, MA, USA). Head isosurfaces were created by adjusting a grey-scale threshold value that optimized skin segmentation for each participant and facial region. The result of this manual segmentation was used to train the automatic head reconstruction algorithms in BioFace3D.

Additionally, three trained researchers manually registered the 3D coordinates of anatomical facial landmarks on the reconstructed surfaces using the Amira software. This landmark selection included homologous landmarks that could be reliably located on the facial structures of all individuals in the sample, and have been used and validated in previous works [6, 37, 38] –see S2 Fig for landmark definition. These manually annotated landmarks were used to train two of the automatic landmarking models included in BioFace3D.

In addition to this proprietary dataset, the development of the head alignment BioFace3D module required training additional facial landmarking models with the 1212 3D facial models of the publicly available LYHM dataset [39].

## BioFace3D framework

BioFace3D is divided into three major modules (Fig 1), comprising: *i)* facial mesh extraction from MRI (Fig 1A), *ii)* 3D facial landmarking (Fig 1B) and *iii)* facial biomarkers computation (Fig 1C). To perform an end-to-end facial biomarker study, it is necessary to iteratively apply modules 1 and 2 to all individuals in the sample. Finally, from the 3D coordinates of the landmarks, the biomarker analysis can be completed with module 3.

Each module has its own user interface (UI). The software was written in MATLAB 2023b [41], with Deep Learning, Statistics and Machine Learning, Image Processing, Computer Vision, and Mapping Toolboxes, and with dependencies on Python 3.8 and TensorFlow [42]. It can be run on a Windows personal computer (16GB RAM with 6-core CPU). The use of GPU can accelerate the execution of some of its components.

## Module 1: Face extraction from MRI

This module automatically produces a facial mesh from an MRI scan, taking as its input NIfTI file(s) (.nii/.nii.gz) and generating 3D facial meshes encoded in PLY file(s) (.ply) as output. It can process files individually or in batch mode.

Fig 2A shows the UI of the facial mesh extraction module. It is designed as a multi-step process that can be interrupted by the user to supervise and validate intermediate results (Fig 2B and Fig 2C). Algorithm 1 presents a pseudo-code of the automatic processes performed by the module, which are detailed below.

**Orientation normalization** (① **in Fig 1**). The orientation of an anatomical structure –such as the head– in an MRI scan is determined by the patient-based coordinate system defined by the DICOM standard [44]. These coordinates relate the Cartesian axes to the anatomical coordinate system. Generally, NIfTI or DICOM files define RAS (x-axis pointing to Right, y-axis to Superior, z-axis to Anterior) as the default orientation. This information is stored in the `DataDirection` field of the NIfTI file header, codifying the reference orientation as a 3x3 diagonal matrix.

However, depending on the reconstruction software, technique used or post-acquisition alignment procedures, it is possible to find up to 24 possible orientations, assuming that the coordinates must always be orthogonal and comply with the right-hand rule [45]. This information is codified in the `QForm` field of the NIfTI file header as a 3x3 rotation matrix with respect to the orientation determined by `DataDirection`. In this way, we can know the orientation of all MRI files and apply the corresponding rotation transformations to automatically orient all heads to a common direction. Following [36], in BioFace3D the orientation of the heads is normalized to the LSA orientation (x-axis pointing to Left, y-axis to Anterior, z-axis to Superior).

**Slice processing & enhancement** (② **in Fig 1**). Most head MRI scans are acquired to analyze the brain, so the surrounding regions often present problems of non-uniform illumination and signal attenuation. To reduce this variability, a harmonization process is performed to reduce intensity inhomogeneity, either intra-MRI or inter-MRI [46]. This processing is divided into two stages: histogram matching and bias field correction (see S3 Fig).

To normalize intensity and reduce variability across MRI slices, histogram matching is applied using as a reference an MRI scan from the IXI dataset [47] that shows high contrast among tissues and uniform intensity across slices. The SimpleITK [48] histogram matching implementation is applied on any input MRI, increasing intensity uniformity. Next, to reduce non-uniform intensity bias, the N4ITK (N4 Inhomogeneity Correction) method is used [49]. To determine the MRI region where bias correction is applied, a binary mask is created by thresholding the histogram-matched MRI, and subsequently applying the hole-filling and closing mathematical morphology operations. The result of the bias field and logarithmic bias correction process are portrayed in S3 Fig.

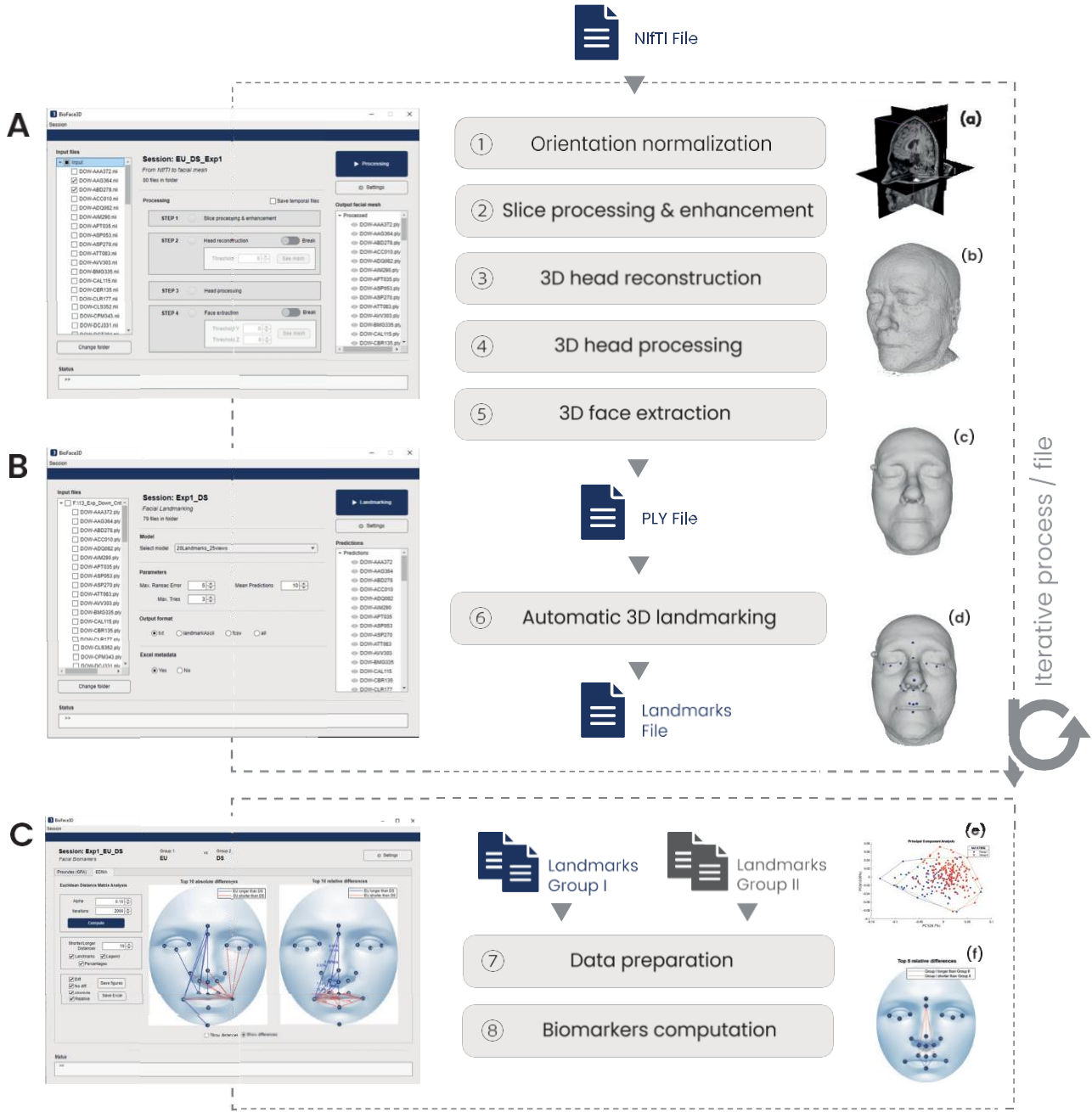

**Fig 1. BioFace3D framework.** (A) Module for MRI processing and 3D facial mesh extraction. ① From an (a) MRI, the orientation of the head is normalized to a common direction. ② Image processing and enhancement to correct for intensity variations along the MRI 2D slices. ③ Estimation of the optimal facial soft tissue segmentation threshold and reconstruction of the (b) 3D head mesh. ④ Processing of the 3D head for centering, emptying and alignment. ⑤ Extraction and remesh of the (c) 3D facial mesh and storage as a PLY file. (B) Module of 3D facial landmarking. From the PLY file, ⑥ an automatic prediction of the (d) 3D coordinates of the landmarks is performed with pre-trained multi-view consensus CNNs [36] models. (C) Module for facial biomarkers computation between groups. ⑦ From the landmark files, data are prepared for biomarker extraction. ⑧ Computation of facial biomarkers based on GM techniques. (e) GPA and (f) EDMA biomarkers identify statistically significant global and local differences in facial morphology between groups.

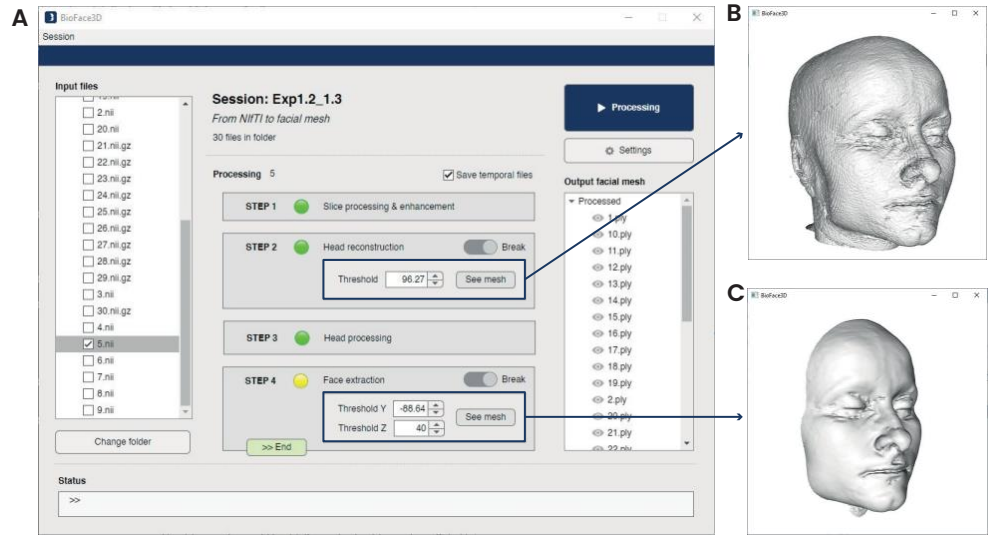

**Fig 2. Module 1: Face extraction from MRI.** (A) UI of the module. The left side panel lists the files in the input folder (.nii/.nii.gz) that can be selected to process, and the right side panel lists the already processed (.ply) files. The central panel shows the different steps that are automatically completed to obtain a 3D facial mesh from the MRI. (B) Intermediate preview of the 3D head reconstruction and manual adjustment of the segmentation threshold. (C) Intermediate preview of the resulting 3D facial mesh and manual adjustment of the cutting thresholds.

**3D head reconstruction (③ in Fig 1).** The 3D head models are reconstructed from the MRI scan by automatically determining a greyscale intensity threshold that optimizes skin segmentation and distinguishes between background and foreground voxels (see Fig 3).

For this purpose, facial anatomy experts were asked to manually select the optimal threshold value for the segmentation of a subset of 150 MRI scans from our training dataset. Then, we measured a correlation coefficient of $r = 0.81$ between the mean and maximum intensity level of the MRI and the manually selected optimal threshold. For this reason, we trained a linear regression model to automatically estimate this segmentation threshold value using the mean and maximum intensity level of the MRI as input variables. Based on this threshold, the Flying Edges algorithm [50] is applied for generating a 3D binary polygonal reconstruction of the head. As shown in Fig 2B and Fig 3, BioFace3D predicts the optimal segmentation threshold. However, if the user is unsatisfied with the results, the predicted threshold value can be adjusted through the UI and the resulting 3D head mesh can be previsualized.

**Head processing (④ in Fig 1).** This is a denoising and normalization process, which is implemented using pymeshlab [51]. It involves four steps (see Fig 4): (1) cleaning of isolated elements, (2) centering of the head bounding box, (3) fine alignment to a reference, and (4) internal emptying of the head.

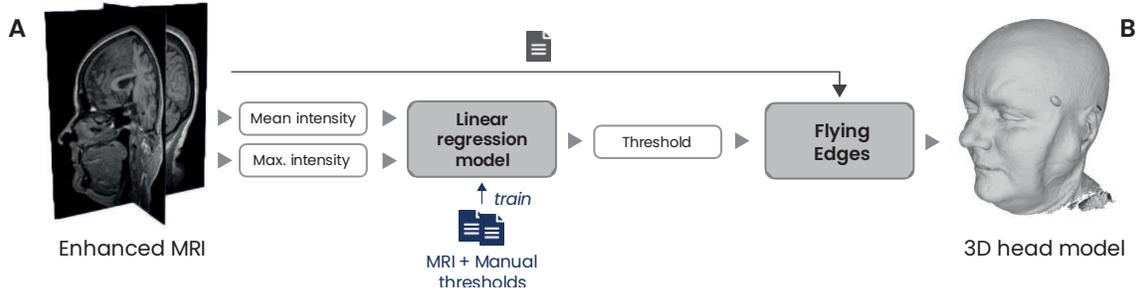

**Fig 3. 3D head reconstruction from MRI.** From the (A) enhanced MRI, a linear regression model predicts the optimal threshold based on the mean and maximum intensity voxel values. (B) 3D head model reconstructed from MRI using the Flying Edges algorithm [50] and the predicted threshold.

The 3D head models resulting from the reconstruction usually present small isolated elements corresponding to the noise outside the head. To remove this noise and keep only the head mesh, a 3D filter is applied to remove isolated elements with a diameter less than 90% of the maximum mesh size –corresponding to the head. This allows to define the 3D bounding box of the head and normalize its position in terms of translation. To do this, we calculate the centroid of the 3D bounding box and use this value to translate the entire mesh to the origin of the Cartesian coordinate system (0,0,0) (see Fig 4B).

To complete the fine alignment of the 3D head, a 3D rigid transformation is applied to align all heads to exactly the same reference (see S4 Fig). To that end, we use a reference head mesh that has benn manually aligned in the LSA orientation, and registered with 5 anatomical control points corresponding to the inner corners of the eyes, the tip of the nose and the corners of the mouth. We refer to the coordinates of these control points as $CP_{ref} \in \mathbb{R}^{5\times 3}$. Subsequently, we automatically register these 5 control points ($CP_{in} \in \mathbb{R}^{5\times 3}$) on the input head mesh to be aligned, leveraging a Multi-view Consensus Convolutional Neural Network (MVCNN) model [36] trained on the LYHM dataset [39]. Next, we perform an inter-point match that estimates the similarity transformation that maps the position of $CP_{in}$ to that of $CP_{ref}$. And then, the corresponding 3D rigid geometric transformation is applied to the input head mesh to perfectly align it to the reference (see Fig 4C). It is important to note that since this may be a time-consuming process, this step can be disabled from the BioFace3D UI.

Finally, to empty the head, internal vertices are automatically selected based on their quality and subsequently removed. To this end, an ambient occlusion filter is applied and vertices are categorized in terms of their quality in the 0 to 1 range. The darkest (or deepest) vertices –corresponding to the most internal areas of the head– will have values close to 0. In contrast, the vertices of the most external facial surface will have quality values close to 1. Finally, vertices with a quality between 0 and 0.25 are automatically selected and removed (see Fig 4D).

**3D facial mesh extraction (⑤ in Fig 1).** Finally, given that the head has been already aligned in a common reference space, we can apply fixed thresholds to cut the 3D mesh along the Z (*thZ*) and Y (*thY*) axes and extract only the facial surface. These threshold values are displayed in the BioFace3D UI and can be modified by the user, offering also a preview of the result, as shown in Fig 2. In addition, all vertices and faces that have ended up isolated due to the breaking the original mesh are also removed.

Optionally, to further improve the visual quality of the resulting 3D facial mesh, a Taubin low-pass filter [52] can be applied to smooth the facial surface and to eliminate the high-frequency noise resulting from the 3D scanning and reconstruction. Moreover, to close possible holes left on the facial surface, the screened Poisson reconstruction method [53] can be applied, which generates a new isosurface that preserves the original morphology of the face. This filter is based on Neumann boundary conditions, so the newly generated mesh extends beyond the boundaries of the original face mesh. Therefore, the same cropping strategy explained above is applied on the mesh resulting from the screened Poisson reconstruction. It is important to note that the Taubin low-pass filtering and screened Poisson reconstruction are optional to the user, since the new meshes may present small variations with respect to the original mesh. These slight differences are discussed in the results section.

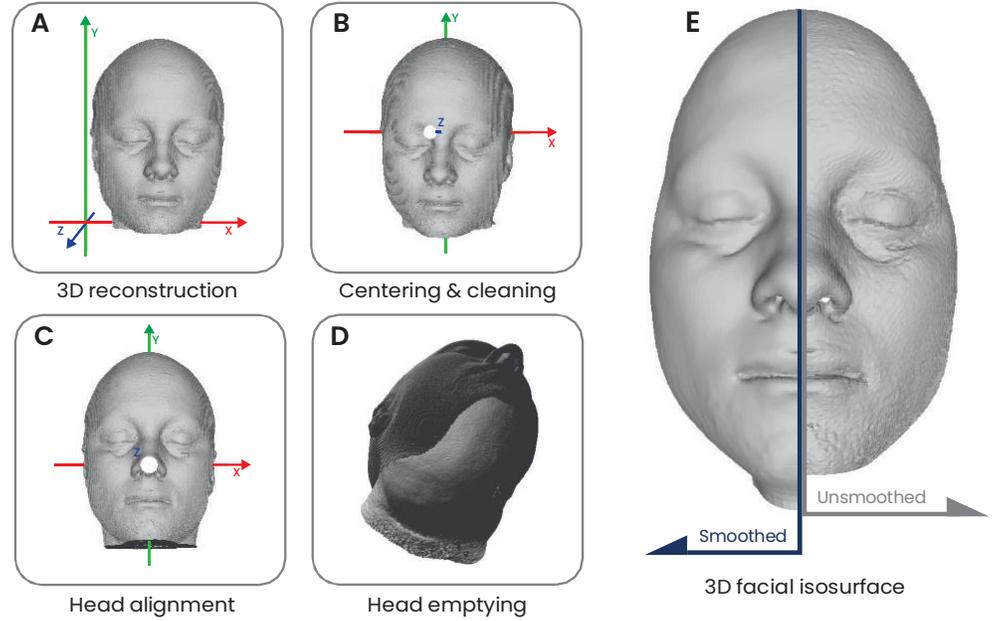

**Fig 4. 3D head processing.** (A) 3D head model reconstructed from MRI. (B) 3D head model after cleaning and centering of the bounding box at the origin of coordinates. (C) 3D head model perfectly aligned. (D) 3D head model internally emptied. (E) 3D facial isosurface extracted from the head before applying a remeshing and smoothing process (right) and after applying it (left).

**Algorithm 1.** 3D facial mesh extraction from MRI

```
Input: Head MRI file M;
Output: 3D facial isosurface F ∈ R^(n×3)
   → Rotate to LSA orientation
1:   orientation_init = GetOrientation(M)
2:   M_reoriented = TransformOrientation(orientation_init, 'LSA')
   → MRI slice processing
3:   M1_enhanced = HistogramMatching(M_reoriented, Template)
4:   mask = GetMask(M1_enhanced)
5:   M2_enhanced = BiasCorrection(M1_enhanced, mask)
   → 3D Head reconstruction
6:   [MaxI, MeanI] = GetIntensity(M2_enhanced)
7:   th = PredictThreshold(MaxI, MeanI)
8:   if th is valid
       → Get a PLY H ∈ R^(n×3) model of the head
```

```
 9:         H = FlyingEdges3D(M2_enhanced, th);
         → 3D head mesh processing
10:         H_clean = RemoveIsolatedPieces(H, diameter = 90%)
11:         [DimX, DimY, DimZ] = GetBoundingBox(H_clean)
12:         H_centered = CentreMesh(H_clean, [DimX, DimY, DimZ])
13:         if fine_alignment is true
            → Fine alignment 3D head
14:             CP_in = MVCNN(H_centered)
15:             H_aligned = SimilarityTransformation(CP_in, CP_ref)
16:         else
                H_aligned = H_centered
17:         end if
         → Emptying mesh
18:         H_ambient = AmbientOcclusion(H_aligned)
19:         vertex_selected = SelectVertexQuality(MinQuality = 0, MaxQuality = 0.25)
20:         H_empty = RemoveSelectedVertex(H_ambient, vertex_selected)
         → 3D facial mesh extraction  T ∈ R^{n×3}
21:         T = ExtractFacialMesh(H_empty, thY, thZ)
22:         if smooth_remesh is true
            → Apply remesh and smoothing filters
23:             T_Taubin = TaubinFilter(T)
24:             T_Poisson = ScreenedPoisson(T_Taubin)
25:             F = T_Poisson
26:         else
27:             F = T
28:         end if
29:         if save_snapshot is true
30:         → Save a 2D snapshot of final        mesh (F)
            facial
31:             SaveSnapshot(F)
32:         end if
33: else
34:     → Set threshold (th) and retake from line 9
35: end if
36: return F
```

## Module 2: 3D landmarking

The 3D landmarking module (⑥ in Fig 1) takes as input data the 3D facial meshes in PLY (.ply) format from the previous module and generates output data file(s) in various formats with the automatically predicted facial 3D landmarks coordinates. Fig 5A presents the UI of the module.

In BioFace3D, automatic 3D facial landmarking is performed using state-of-the-art Multi-view Consensus Convolutional Neural Network (MVCNN) [36]. This method is based on performing a 2D projection of the 3D head models from different viewpoints, using the hourglass model to predict candidate landmarks on 2D heat maps and ray-projecting the final position onto a real vertex of the 3D mesh by using least squares (LSQ) combined with random sample consensus (RANSAC). This 2D multi-view approach allows 3D landmark registration at a reduced computational cost but with high precision. The main training hyperparameters affecting the model are the number of views (i.e. number of 2D projections with random camera positions taken) and the image channels.

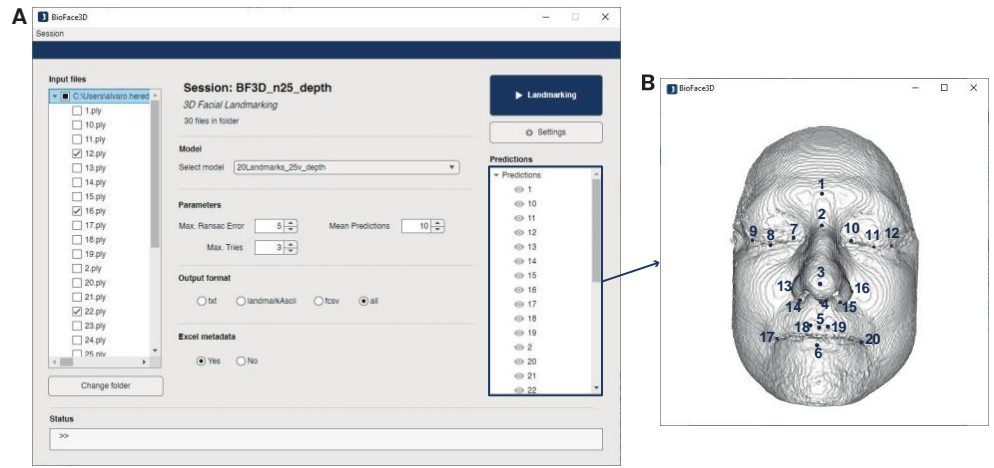

**Fig 5. Module 2: 3D landmarking.** (A) UI of the module. The left side panel lists the files in the input folder (.ply) that can be selected to landmark, and the right side panel lists the already landmarked files. In the central panel, it is possible to select the pre-trained MVCNN model to perform landmarking, modify prediction hyperparameters and adjust the output formats. (B) Previsualization of landmark prediction from UI using BioFace3D-20 landmark model.

For flexibility, BioFace3D includes two pre-trained MVCNN models for facial landmarking, so the user chooses the best suited model.

- **BioFace3D-20 model**: 20 landmarks, trained on 538 3D facial models segmented from MRI scans. Training hyperparamters: 25 views, depth + geometry channels. See S2 Fig for landmarks definition.

- **DTU-3D model**: 73 landmarks (24 landmarks + 49 semi-landmarks), trained on 601 3D facial models obtained with a Canfield Vectra M3 surface scanner. See [36] for landmarks definition and training hyperparameter details.

### Module 3: Facial biomarkers

This last module allows the statistical comparison between two groups of individuals (i.e. patients and healthy subjects) through the automatic calculation of biomarkers. These biomarkers, obtained from their homologous landmarks by GM, characterize the diversity of faces in vectors that encode their morphology. Fig 6 shows the different sections of the UI module.

**Data preparation** (⑦ in Fig 1). Prior to the calculation of biomarkers, a data preparation process is required. From the landmark files of the previous module, it is necessary to generate a spreadsheet for each study group with the coordinates of the individuals' landmarks. This process can be completed automatically from the UI.

**Biomarkers computation** (⑧ in Fig 1). Two main biomarker calculation techniques have been implemented: Generalized Procrustes Analysis (GPA) and Euclidean Distance Matrix Analysis (EDMA).

- **GPA biomarkers (Fig 6A).** To obtain GPA-based facial biomarkers, it is necessary to remove the influence of scale and position from the landmarks configurations, so that they are in the same morphospace [19]. For this, the coordinates of the landmarks of each subject are normalized by their centroid size (defined as the square root of the sum of squared distances of all the landmarks to their centroid), translated to the same origin of coordinates, and rotated until achieving a least-squares fit between homologous landmarks, minimizing the Procrustes distance. Then, the Procrustes coordinates of each landmark configuration can be interpreted as a single $n$-dimensional vector, where each dimension is equivalent to a spatial coordinate of one of the landmarks. Considering that we work with $l$ 3D landmarks, vectors will have a dimension of $n = 3 \cdot l$.

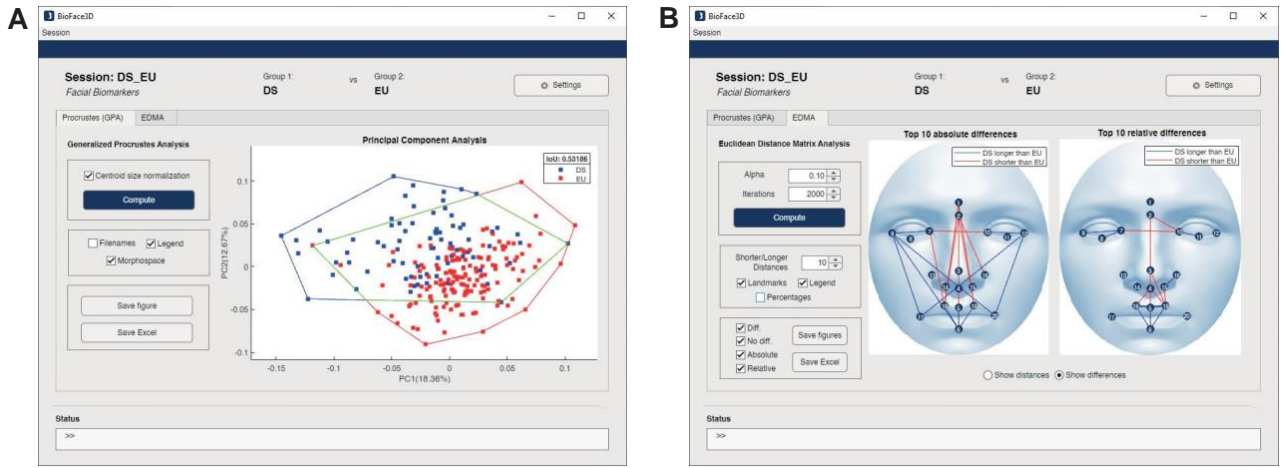

**Fig 6. Module 3: Facial biomarkers.** UI of the module is divided into two sections: (A) GPA analysis and (B) EDMA analysis. Each section allows the configuration of analysis parameters and the saving of figures and metadata files.

This set of vectors is then decomposed using Principal Component Analysis (PCA) into its principal components, i.e. linear combinations of the original dimensions that simplify the description of the variation between individuals in the sample and allow the visualization of natural groupings of individuals in spaces of reduced dimension. In this way, it is easier to analyze which groups of individuals have a more similar facial shape. GPA-based facial biomarkers are the first vector components obtained after the dimension reduction of PCA. Therefore, different facial biomarkers can be built based on GPA retaining a greater or lesser number of principal components. As user feedback, the UI of this module displays the morphospaces generated from the two principal components. In addition, the convex hull of each morphospace can be displayed, as well as the intersection over union (IoU) metric as a measure of the separability of the two groups.

- **EDMA biomarkers (Fig 6B).** These are based on the calculation of the matrix of linear distances between all possible pairs of landmarks of each individual. This enables to statistically determine which linear measurements differ between the two study groups by pairwise comparisons based on confidence interval (α, commonly α = 10) tests [22].

  This analysis is performed under the statistical hypothesis that for the two groups and given a specific inter-landmark distance, this has the same expected mean value for both groups. If the hypothesis is proven true, this means that inter-landmark distance is not significantly different between the two groups. After all the statistical tests have been carried out on each of the inter-landmark distances, those that are found to be statistically different between both groups will be considered the distinctive features of one group with respect to the other.

These characteristic distances are ordered according to the degree of difference, from greatest to least. Therefore, we can construct different facial biomarkers based on EDMA using a greater or lesser number of inter-landmark distances among the most significant differences between both groups.

In the UI, the number of significantly different inter-landmark distances used for the construction of the EDMA facial biomarkers can be selected from 2 to the number of significantly different distances obtained in each experiment. In addition, the percentage of significantly different distances is indicated. These distances are iteratively displayed on a simulated facial figure with the landmarks (see Fig 6B).

## Conclusions

This work has introduced BioFace3D, an end-to-end automatic pipeline for computing GM-based facial biomarkers of diagnostic potential on 3D facial models extracted from structural MRI scans of the head. As such, the presented pipeline enables high-throughput processing that starts from NIfTI MRI scan files and finishes with GM facial biomarker-based comparison between diagnostic groups. While the whole processing is completely automatic, BioFace3D allows user intervention at different points of the process to adjust and refine intermediate results, thus ensuring the best accuracy. In the future, we will extend BioFace3D in the following directions: *i)* the adaptation of Module 1 to handle other input 3D facial imaging modalities, like stereophotogrammetry 3D facial models, and *ii)* the inclusion of machine learning models that can be trained on the extracted facial biomarkers to perform diagnostic classification from facial shape data.

## Supporting information

**S1 Table. Training datasets description.** Formed by 538 T1w head MRIs from different centres and scanner parameters, with representation of individuals of both sexes.

**S2 Fig. Landmarks definition.** Location and definition of BioFace-20 landmarks model.

**S3 Fig. MRI slice processing steps of a single middle sagittal slice.** (A) Original sagittal slice, (B) histogram-matched enhanced slice, (C) bias-corrected sagittal slice, (D) IXI template used for histogram matching, (E) matched histogram, (F) binary mask used for bias field correction and (G) log bias.

**S4 Fig. Fine alignment process by matching reference control points.** (A) 5-landmarks selection of LYHM dataset, (B) example of a head before and after applying fine alignment, (C) example of $CP_{in}$ and $CP_{ref}$ with corresponding transformation and (D) transformation applied for alignment from different viewpoints.

## Acknowledgments

The research in this paper was supported by the Joan Oró grant (2024 FI-3 00160) from the Recerca i Universitats Departament (DRU) of the Generalitat de Catalunya with grant 2023 FI-2 00160 and the European Social Fund, by Agencia Española de Investigación (PID2020-113609RB-C21/AEI/10.13039/501100011033), by Instituto de Salud Carlos III (ISCIII) through the contracts FI21/00093 and CP20/00072 (co-funded by European Regional Development Fund (ERDF)/European Social Fund "Investing in your future") and by Fondation Jerome Lejeune with grant 2020b cycle-Project No.2001. The authors would also like to thank the Agència de Gestió d'Ajuts Universitaris i de Recerca (AGAUR) of the Generalitat de Catalunya (2021 SGR01396, 2021 SGR00706, 2021 SGR1475).

25. Li M, Cole JB, Manyama M, Larson JR, Liberton DK, Riccardi SL, et al. Rapid automated landmarking for morphometric analysis of three-dimensional facial scans. J Anat. 2017;230: 607–618. https://doi.org/10.1111/joa.12576

26. Ramon E, Triginer G, Escur J, Pumarola A, Garcia J, Giró-i-Nieto X, et al. H3D-Net: Few-Shot High-Fidelity 3D Head Reconstruction. 2021 IEEE/CVF International Conference on Computer Vision (ICCV). 2021. pp. 5600–5609. https://doi.org/10.1109/ICCV48922.2021.00557

27. Liu Y, Li L, An S, Helmholz P, Palmer R, Baynam G. 3D Face Reconstruction with Mobile Phone Cameras for Rare Disease Diagnosis. In: Aziz H, Correˆa D, French T, editors. AI 2022: Advances in Artificial Intelligence. Cham: Springer International Publishing; 2022. pp. 544–556. https://doi.org/10.1007/978-3-031-22695-3_38

28. Hammond P, Suttie M. Large-scale objective phenotyping of 3D facial morphology. Hum Mutat. 2012;33: 817–825. https://doi.org/10.1002/humu.22054

29. Marcucio R, Hallgrı́msson B, Young NM. Facial Morphogenesis: Physical and Molecular Interactions Between the Brain and the Face. Current topics in developmental biology. 2015;115, 299–320. https://doi.org/10.1016/bs.ctdb.2015.09.001

30. Hamadelseed O, Chan MK, Skutella T. Distinct neuroanatomical and neuropsychological features of Down syndrome compared to related neurodevelopmental disorders: a systematic review. Frontiers in Neuroscience. 2023;17, 1225228. https://doi.org/10.3389/fnins.2023.1225228

31. Statistical Parametric Mapping: The Analysis of Functional Brain Images

32. Reuter, M., Schmansky, N.J., Rosas, H.D., Fischl, B. Within-Subject Template Estimation for Unbiased Longitudinal Image Analysis. Neuroimage. 2012;61 (4), 1402-1418. https://doi.org/10.1016/j.neuroimage.2012.02.084

33. Berends B, Bielevelt F, Schreurs R, Vinayahalingam S, Maal T, de Jong G. Fully automated landmarking and facial segmentation on 3D photographs. Sci. Rep. 2024;14, 6463. https://doi.org/10.1038/s41598-024-56956-9

34. Wang H, Fang S. Geometric Analysis of 3D Facial Image Data: A Survey. Recent Patents on Engineering. 2022;16: 36–48. https://doi.org/10.2174/1872212116666220117125432

35. Lagravère MO, Low C, Flores-Mir C, Chung R, Carey JP, Heo G, et al. Intraexaminer and interexaminer reliabilities of landmark identification on digitized lateral cephalograms and formatted 3-dimensional cone-beam computerized tomography images. Am J Orthod Dentofacial Orthop. 2010;137: 598–604. https://doi.org/10.1016/j.ajodo.2008.07.018

36. Paulsen RR, Juhl KA, Haspang TM, Hansen T, Ganz M, Einarsson G. Multi-view consensus CNN for 3D facial landmark placement. 201911361: 706–719. https://doi.org/10.1007/978-3-030-20887-5_44

37. Starbuck JM, Llambrich S, González R, Albaigès J, Sarlé A, Wouters J, et al. Green tea extracts containing epigallocatechin-3-gallate modulate facial development in Down syndrome. Sci Rep. 2021;11: 4715. https://doi.org/10.1038/s41598-021-83757-1
October 1, 2024                                                                                       15/17